\documentclass{article}

\PassOptionsToPackage{numbers,compress}{natbib}

\usepackage[preprint]{neurips_2026}
\makeatletter
\renewcommand{\@noticestring}{Preprint. \today.}
\makeatother

\usepackage[utf8]{inputenc}
\usepackage[T1]{fontenc}
\usepackage[hidelinks]{hyperref}
\usepackage{url}
\usepackage{booktabs}
\usepackage{amsmath,amssymb,amsfonts}
\usepackage{nicefrac}
\usepackage{microtype}
\usepackage{xcolor}
\usepackage{graphicx}
\usepackage{enumitem}
\usepackage{tikz}
\usetikzlibrary{positioning,arrows.meta,shapes.geometric,fit,calc,backgrounds,decorations.pathreplacing}
\usepackage{placeins}

\title{Graph Computation Meets Circuit Algebra:\\A Task-Aligned Analysis of Graph Neural Networks for Electronic Design Automation}

\author{%
  Hyunmog Kim\thanks{Independent Researcher. Correspondence to: Hyunmog Kim \texttt{<hyunmog2020@kaist.ac.kr>}.}
}

\begin{document}
\maketitle

\begin{abstract}
EDA problems are graph-structured, but not all graph-structured problems call for the same GNN computation. We argue that successful GNN-for-EDA methods are those whose propagation, aggregation, and supervision align with the native algebra of the target task. Concretely: static timing analysis is a max-plus / min-plus recurrence on a topologically ordered DAG, structurally aligned with asynchronous DAG-GNNs; placement is governed by hypergraph wirelength and density penalties and is exploited by differentiable placers rather than by message-passing GNNs alone; routing congestion is a sparse demand-supply field over a layout grid; switching-activity propagation is a probabilistic recurrence on a directed netlist; IR drop is a linear system on the power-delivery network; and analog symmetry extraction is a discrete constraint-prediction problem on schematic graphs. Through these task-by-task alignments we (i) review the GNN architectural toolkit relevant to circuits, (ii) formalize how circuit graphs differ from generic graphs (directed, heterogeneous, multi-scale, with sequential and clock structure), (iii) characterize where current methods succeed and where the algebra--architecture mismatch limits them, and (iv) identify failure modes---stage leakage, proxy-to-signoff gap, calibration, and design-distribution shift---that we believe are likely to dominate the next phase of work. We position the paper as a GNN-for-EDA, task-aligned analysis rather than a comprehensive AI-for-chip-design survey. Continuous SE(3)-equivariant geometric GNNs are usually mismatched to Manhattan digital layout, and LLM-for-RTL, HLS, and RL/diffusion-based topology generation are outside our scope.
\end{abstract}

\section{Introduction}
\label{sec:intro}

Modern semiconductor design is at a difficult inflection. The slowdown of Dennard scaling has not reduced demand for new chips---if anything, AI accelerators, automotive SoCs, and chiplet-based systems have accelerated it---but it has dramatically raised the cost of physical realization. Designs at advanced nodes routinely contain tens to hundreds of millions of placeable instances, dozens of routing layers, and intricate constraints on power, timing, signal integrity, and manufacturability. Place-and-route iterations frequently take days, and full sign-off-quality runs can take weeks. Within this loop, design teams must explore enormous configuration spaces under tight schedules, and the cost of late-stage iteration motivates earlier, faster predictions.

Graph neural networks (GNNs) have repeatedly been proposed as a tool for this loop. The argument is appealing: circuits \emph{are} graphs---a netlist is a hypergraph of cells and nets, a placement annotates that graph with coordinates, a clock tree is a directed tree, and a 2.5D chiplet system is a hierarchical graph of dies, interposers, and through-silicon vias. Flat-vector and image-based models throw away the relational structure that experts use to reason about timing paths, fan-out cones, and buffer trees. GNNs, in principle, were designed for this kind of reasoning.

\paragraph{Thesis.} Yet success has been uneven. Some GNN-for-EDA results---Google's AlphaChip as a publicly reported production floorplanning case, and NVIDIA's GRANNITE as an industrial-authored power-estimation result evaluated against commercial-flow references---have made a credible case for stage-specific industrial relevance, while many academic claims have been hard to reproduce or have failed to generalize across designs and technology nodes. We argue that the difference is rarely about the message-passing variant chosen. Rather:
\begin{quote}
\itshape Successful GNN-for-EDA methods are not defined by using message passing; they are defined by matching the graph computation to the algebra of the target EDA task.
\end{quote}
Concretely: static timing analysis (STA) is a max-plus / min-plus recurrence on a directed acyclic graph (DAG); placement is the optimization of a hypergraph wirelength functional under a density constraint; congestion is a sparse demand-supply field; power is a probabilistic recurrence; IR drop is a linear system on a Laplacian; and analog symmetry extraction is a discrete constraint-prediction problem. Each task has a native algebraic shape, and useful GNNs are those whose forward computation respects that shape.

\paragraph{Contributions.} This paper makes four contributions:
\begin{enumerate}[noitemsep,leftmargin=*]
    \item A compact, EDA-relevant tour of the GNN architectural toolkit (\S\ref{sec:toolkit}), emphasizing the heterogeneous, hypergraph, and topological-DAG variants that recent EDA papers actually use.
    \item A formal characterization of \emph{circuit graphs} as multi-typed, partially ordered, multi-scale objects (\S\ref{sec:circuit_graphs}), with explicit caveats on where the DAG / homogeneous / clique-expansion assumptions of generic graph models break down.
    \item A task-by-task algebraic analysis (\S\ref{sec:applications}) connecting each major EDA stage---logic synthesis, placement, congestion, timing, power and IR drop, analog symmetry---to graph-computation primitives aligned with its mathematical structure, using explicit task notation and calibrated industrial-evidence claims.
    \item A characterization of the open challenges that, in our view, will determine whether GNNs become a durable fixture of EDA flows: \emph{stage leakage} (using post-route features for pre-route prediction), the \emph{proxy-to-signoff gap}, calibration and uncertainty for high-cost wrong predictions, and design-distribution shift across PDK / library / tool (\S\ref{sec:open}).
\end{enumerate}

\paragraph{Scope.} This paper is not a comprehensive survey of AI for chip design: LLMs for RTL, high-level synthesis, RL/diffusion topology generation, and EDA agent workflows are out of scope, as are non-GNN ML predictors that do not define a graph, except as baselines or flow context. We position the paper as a \emph{GNN-for-EDA, task-aligned analysis} (inclusion criteria in Appendix~\ref{app:methodology}); analog appears only via the discrete symmetry-constraint extraction line. Broad surveys of GNNs and of their real-world challenges~\cite{wu2021comprehensive,ju2025realworld} provide complementary coverage. We deliberately do not cover continuous SE(3)/E(3)-equivariant geometric GNNs (SchNet, EGNN, MACE, Equiformer)~\cite{schutt2017schnet,satorras2021egnn,batatia2022mace,liao2024equiformerv2}: digital layout obeys a Manhattan grid with discrete cell orientations and direction-preferred metal layers, breaking the continuous rotational symmetry these architectures encode. Where 2D symmetry does exist---in lithographic hotspot patterns or analog matched-device pairs---the relevant group is a finite subgroup of $E(2)$ (typically $C_4$ or $D_4$), better captured by 2D group-equivariant CNNs or by augmentation over the discrete legal orientation set than by point-cloud equivariant message passing. Weaker symmetries (translation invariance, reflections, analog matching constraints) remain relevant and are discussed where they appear.

\section{Preliminaries}
\label{sec:prelim}

A graph $\mathcal{G} = (\mathcal{V}, \mathcal{E})$ has nodes $\mathcal{V}$ ($|\mathcal{V}|=n$) and edges $\mathcal{E}$ ($|\mathcal{E}|=m$). Nodes carry feature vectors $\mathbf{x}_v \in \mathbb{R}^d$; edges may carry attributes $\mathbf{e}_{uv} \in \mathbb{R}^c$. Three structural variants are central for circuits: \emph{directed} graphs (driver-to-load signal flow), \emph{heterogeneous} graphs (cells, nets, pins; signal, clock, power as distinct types), and \emph{hierarchical} graphs (chiplet super-nodes over within-die netlists).

The standard message-passing primitive~\cite{gilmer2017} is
\begin{equation}
\label{eq:mpnn}
\mathbf{h}_v^{(\ell+1)} = \phi\!\left(\mathbf{h}_v^{(\ell)},\; \mathrm{AGG}\!\left(\{\mathbf{h}_u^{(\ell)} : u \in \mathcal{N}(v)\}\right)\right),
\end{equation}
where $\mathrm{AGG}$ is permutation-invariant. After $K$ rounds, every node aggregates information from its $K$-hop neighborhood. Choices for $\mathrm{AGG}$ (sum, mean, attention) and for whether $\mathcal{N}(v)$ respects edge type or direction define the architectural landscape we review next.

\section{GNN Toolkit for Circuits}
\label{sec:toolkit}

We summarize only the architectural ideas that are actually load-bearing for EDA work. A more detailed review, including spectral foundations, GAT/GIN derivations, and theoretical pathologies, is in Appendix~\ref{app:gnn_review}.

\paragraph{Spatial message passing.} GCN~\cite{kipf2017} performs degree-normalized neighbor aggregation; GraphSAGE~\cite{hamilton2017} introduces inductive sampling crucial for circuit-scale graphs (full-batch GCN does not fit on a single GPU at industrial scale). GAT~\cite{velickovic2018,brody2022gatv2} replaces fixed weights with learned attention---attractive in EDA because different fan-in sources, timing-critical arcs, or typed relations may deserve different weights. GIN~\cite{xu2019gin} gives the maximum expressive power of standard message passing by combining sum aggregation with an MLP update, and matches the 1-WL graph isomorphism test as an upper bound.

\paragraph{Heterogeneous and relational variants.} Real netlists are not type-homogeneous, and conflating types loses essential structure. R-GCN~\cite{schlichtkrull2018rgcn} applies relation-specific weight matrices,
\begin{equation}
\label{eq:rgcn}
\mathbf{h}_v^{(\ell+1)} = \sigma\!\left(\sum_{r \in \mathcal{R}}\sum_{u \in \mathcal{N}_r(v)} \tfrac{1}{c_{v,r}}\, \mathbf{W}_r^{(\ell)}\mathbf{h}_u^{(\ell)} + \mathbf{W}_0^{(\ell)}\mathbf{h}_v^{(\ell)}\right),
\end{equation}
with basis decomposition for parameter sharing. The Heterogeneous Graph Transformer (HGT)~\cite{hu2020hgt} replaces these typed weights with relation-specific Q/K/V projections, giving data-adaptive aggregation at the cost of more parameters. EDA-specific instantiations distinguish at minimum driver-pin$\to$net, net$\to$load-pin, and pin$\to$cell relations on the signal-flow graph (Fig.~\ref{fig:hetero_graph}); clock and power-delivery connections are typically modeled on \emph{separate} graphs (clock tree, PDN), since their physics and constraints differ from signal nets.

\paragraph{Topological / asynchronous DAG message passing.} The most consequential modeling choice for many EDA tasks is whether to respect topological order. Standard MPNNs (Eq.~\ref{eq:mpnn}) update every node in parallel, regardless of position. An alternative respects the topological order $\pi$ of the DAG: each node is visited once, in order, aggregating from predecessors that have already been visited:
\begin{equation}
\label{eq:dag_async}
\mathbf{h}_v = \phi\!\left(\mathbf{x}_v,\; \mathrm{AGG}\!\left(\{\mathbf{h}_u : u \in \mathcal{N}^-(v)\}\right)\right), \qquad v \text{ visited in order } \pi.
\end{equation}
A single forward pass makes information from the entire transitive fan-in cone \emph{reachable} at every node, though it is still compressed into a finite-width hidden state and is therefore not immune to over-squashing. D-VAE~\cite{zhang2019dvae}, DAGNN~\cite{thost2021dagnn}, DeepGate~\cite{li2022deepgate}, and PreRoutGNN~\cite{zhong2024preroutgnn} all adopt this asynchronous variant. The reason is algebraic: arrival time, signal probability, and AIG functional simulation are themselves topological recurrences (\S\ref{sec:applications}), and synchronous GNNs must approximate them with a fixed-depth stack of layers, while an async DAG-GNN aligns with the recurrence ordering---though whether the learned aggregation reproduces the underlying max/min algebra exactly is a separate question of operator and hidden-state capacity. Appendix~\ref{app:gnn_review} contrasts the two regimes graphically.

\paragraph{Hypergraph nets.} A multi-pin net is naturally a \emph{hyperedge} $e \subseteq \mathcal{V}$. Most EDA-GNN papers reduce nets to ordinary graphs by one of three transformations: \emph{clique expansion} (every pin pair becomes an edge, inflating high-fanout nets), \emph{star expansion} (introduce a virtual node per net), or \emph{bipartite cell-net incidence} (pins or cells on one side, nets on the other). Each has different bias: clique expansion overweights high-fanout nets unless edge weights are normalized; star and bipartite formulations preserve the multi-way structure but require careful attribute encoding (pin offset, net degree, routing context) to recover pairwise geometric information~\cite{feng2019hgnn,wang2022lhnn}. Hierarchical pooling methods (DiffPool~\cite{ying2018diffpool}, SAGPool~\cite{lee2019sagpool}) developed in the broader graph-ML literature are less directly used in EDA but motivate the multi-scale formulations of \S\ref{sec:circuit_graphs}.

\paragraph{Graph transformers and pathologies.} Graph transformers (Graphormer~\cite{ying2021graphormer}, GraphGPS~\cite{rampasek2022gps}) bypass local message passing for global attention; scalable variants (NodeFormer~\cite{wu2022nodeformer}, Exphormer~\cite{shirzad2023exphormer}, SGFormer~\cite{wu2023sgformer}, Polynormer~\cite{deng2024polynormer}) attack the $O(n^2)$ cost. Their evidence in graph-ML benchmarks is strong, but their evidence on full-chip industrial EDA is much thinner. Three pathologies---the 1-WL expressiveness ceiling~\cite{xu2019gin}, with higher-order extensions (k-WL GNNs~\cite{morris2019kwl}, PPGN~\cite{maron2019ppgn}) at $O(n^k)$ cost, over-smoothing~\cite{li2018oversmoothing,oono2020oversmoothing} (with mitigations such as JKNet~\cite{xu2018jknet}, GCNII~\cite{chen2020gcnii}, PairNorm~\cite{zhao2020pairnorm}, DropEdge~\cite{rong2020dropedge}), and over-squashing~\cite{alon2021oversquashing,topping2022oversquashing}---constrain depth and reach. In EDA practice, augmenting node features with structural descriptors (degree, hop distance to PIs/POs, position in clock tree) often raises effective expressiveness more cheaply than algorithmic upgrades; over-smoothing is often less visible at the shallow $K \in [2,8]$ depths commonly used in EDA papers, while over-squashing remains salient and argues for graph rewiring or transformer-style global attention when long-range circuit phenomena (global timing, IR drop) matter.

\section{Circuits as Graphs: Representation Choices}
\label{sec:circuit_graphs}

How a circuit is encoded as a graph profoundly shapes what a GNN can learn from it.

\paragraph{Combinational DAG vs.\ full netlist.} A combinational fan-in cone, with sequential elements cut at flip-flop boundaries, is naturally a DAG: nodes are gates, edges are driver-to-load wires, no cycles. \emph{Full} chip netlists, however, contain sequential feedback through state elements, scan logic, latches, generated clocks, and clock-gating cells; treating the entire design as a DAG is incorrect. A robust convention is to apply DAG-based message passing to a \emph{task-specific subgraph}---a timing combinational cone, an AIG fragment, a single clock domain---rather than to the entire physical design. Recent work specifically targets the sequential case: DeepSeq2~\cite{khan2025deepseq2} disentangles structural, functional, and sequential-behavior embeddings rather than collapsing them into a single representation.

\paragraph{And-Inverter Graphs.} In logic synthesis, AIGs offer a normalized representation amenable to functional reasoning and rewriting. Each internal node is an AND of two literals with input-polarity bits, supporting structural hashing and functional fingerprinting. AIGs are the canonical input for the DeepGate line~\cite{li2022deepgate,shi2023deepgate2,shi2024deepgate3} and for symbolic-reasoning approaches such as Gamora~\cite{wu2023gamora}.

\begin{figure}[t]
\centering
\resizebox{0.92\textwidth}{!}{%
\begin{tikzpicture}[
  every node/.style={font=\small},
  cell/.style={rectangle,draw,thick,fill=blue!10,rounded corners=1pt,minimum width=14mm,minimum height=9mm,align=center},
  net/.style={diamond,draw,thick,fill=orange!20,minimum size=9mm,inner sep=1pt,align=center},
  pin/.style={circle,draw,thick,fill=green!35,minimum size=4mm,inner sep=0pt},
  edgelbl/.style={font=\scriptsize,inner sep=1pt},
  rcontain/.style={dashed,semithick,gray!70!black},
  rdrive/.style={->,>=Stealth,thick,blue!70!black},
  rload/.style={->,>=Stealth,thick,red!70!black},
]
\node[cell] (CA) at (0.0, 0) {Cell A};
\node[pin]  (PAo) at (1.5, 0) {};
\node[net]  (N1) at (4.0, 0) {N1};
\node[pin]  (PBi) at (6.5, 0.8) {};
\node[pin]  (PCi) at (6.5,-0.8) {};
\node[cell] (CB) at (8.4, 0.8) {Cell B};
\node[cell] (CC) at (8.4,-0.8) {Cell C};

\draw[rcontain] (CA) -- (PAo);
\draw[rcontain] (CB) -- (PBi);
\draw[rcontain] (CC) -- (PCi);

\draw[rdrive] (PAo) -- (N1) node[midway,above,edgelbl,blue!70!black] {driver};
\draw[rload]  (N1) -- (PBi) node[midway,above=0.5mm,sloped,edgelbl,red!70!black] {load};
\draw[rload]  (N1) -- (PCi) node[midway,below=0.5mm,sloped,edgelbl,red!70!black] {load};

\begin{scope}[shift={(0,-1.7)}]
\node[cell,minimum width=10mm,minimum height=5mm] (lc) at (0,0) {};
\node[right=1mm of lc,font=\scriptsize] (lc_t) {= Cell};
\node[net,minimum size=6mm,right=8mm of lc_t] (ln) {};
\node[right=1mm of ln,font=\scriptsize] (ln_t) {= Net};
\node[pin,right=8mm of ln_t] (lp) {};
\node[right=1mm of lp,font=\scriptsize] (lp_t) {= Pin};
\draw[rcontain] ($(lp_t.east)+(8mm,0)$) -- ++(8mm,0) coordinate (cn_e);
\node[right=1mm of cn_e,font=\scriptsize] {contains};
\end{scope}
\end{tikzpicture}%
}
\caption{A minimal heterogeneous \emph{signal} graph with typed nodes (cells, nets, pins) and typed edges (driver, load, containment): a driver cell exposes a driver pin onto a net that fans out to load pins of downstream cells. Industrial graphs add clock-tree, PDN, hierarchy, physical coordinates, and timing-arc edges, typically modeled on \emph{separate} graphs rather than overloaded onto the signal graph.}
\label{fig:hetero_graph}
\end{figure}
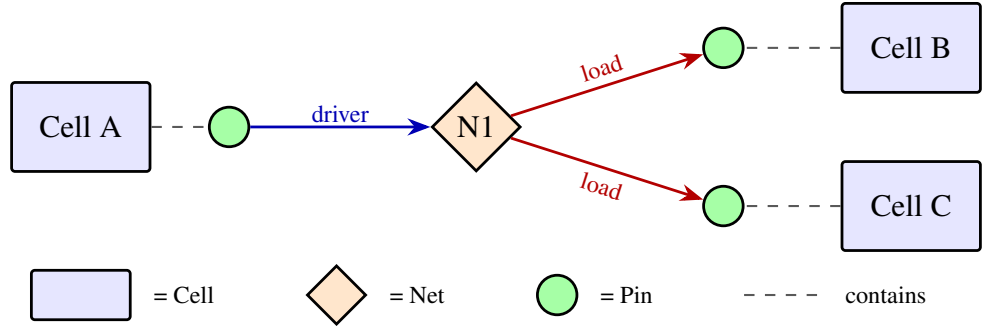

\paragraph{Heterogeneous cell-net-pin graphs.} For physical-design tasks, a uniform "cells as nodes, wires as edges" view loses important structure. Modern formulations distinguish at least three node types---cells (logic gates, macros, flip-flops), nets, and pins---with edge types for driver pin$\to$net, net$\to$load pin, and pin$\to$cell containment (Fig.~\ref{fig:hetero_graph}). R-GCN-style relation-specific weights or HGT-style typed attention treat these distinctions as first-class.

\paragraph{Standard-cell heterogeneity.} Modern libraries are themselves a source of heterogeneity. Mixed-cell-height libraries place single- and multi-row cells together~\cite{jeong2024placement_legalization}, breaking the simplifying assumption of a uniform row pitch. Multi-bit flip-flop (MBFF) cells share clock buffers across bits to reduce power, and come in irregular geometries including L- and T-shaped layouts whose footprint is not captured by a bounding box~\cite{jeong2024dtco_mbff}. Capturing these correctly demands node features beyond a single $(w, h)$ tuple: occupancy masks over standard rows and tracks, per-side pin-access flags, row-span, and orientation-dependent legality. GNNs that ignore them risk misrepresenting L- and T-shaped MBFFs and their downstream impact on clock-tree and routing closure.

\paragraph{Hierarchical and multi-die graphs.} 2.5D and 3D integration introduces a second graph level: chiplets are super-nodes containing within-die netlist sub-graphs; interposer routes and through-silicon vias (TSVs) are inter-chiplet edges. Heterogeneous GNNs in principle reason simultaneously at within-die scale (logic, cell-level timing) and inter-die scale (signal integrity, thermal coupling). Published GNN-for-EDA work that addresses this regime explicitly is still limited; multi-die EDA flows such as Pin-3D~\cite{pentapati2023pin3d} are a context for future learned models, not themselves GNN methods. Appendix~\ref{app:multidie} sketches a more complete multi-scale taxonomy.

\section{Task-Aligned GNN Methods Across EDA Tasks}
\label{sec:applications}

We now connect each EDA stage to the algebraic structure of its target and the GNN computation that matches it. Table~\ref{tab:eda_apps} summarizes representative methods, separating GNN models from non-GNN optimization baselines and from EDA flow contexts.

\subsection{Logic Synthesis}
Logic synthesis transforms an RTL description into an optimized gate-level netlist. The optimization---rewriting, restructuring, technology mapping---is combinatorial; the broader use of GNNs to learn variable-selection policies for branch-and-bound MILP solvers~\cite{gasse2019milp} is a natural conceptual neighbor. The relevant algebra is Boolean rather than metric: useful embeddings should preserve functional equivalence, polarity, reconvergent structure, and simulation-like propagation, not merely local graph neighborhoods. DeepGate~\cite{li2022deepgate} introduced learned AIG embeddings supervised by simulated signal probabilities, yielding a \emph{functional} representation alongside a structural one. DeepGate2~\cite{shi2023deepgate2} sharpens this with pairwise truth-table-difference supervision and a scalable functionality-aware loss; DeepGate3~\cite{shi2024deepgate3} adds Transformer-style attention over the AIG and additional structural objectives, scaling toward larger circuits. Gamora~\cite{wu2023gamora} extends graph-based symbolic reasoning to recover high-level operator structure (adders, multipliers) from gate-level AIGs; PolarGate~\cite{liu2024polargate} explicitly addresses the polarity-aware aggregation step. MGVGA~\cite{wu2025mgvga} is particularly aligned with our thesis: it observes that generic graph masking is algebraically unsafe for circuits because it can break logical equivalence, and instead constrains pretraining through masked gate modeling and Verilog-AIG alignment. Sequential circuits, where state-element feedback is essential, are addressed by DeepSeq2~\cite{khan2025deepseq2}, which disentangles structure, function, and sequential behavior. For AIG-based functional embeddings, the recurrence structure of simulation makes asynchronous DAG message passing (Eq.~\ref{eq:dag_async}) the natural primitive.

\subsection{Placement and Floorplanning}
Placement decides cell coordinates subject to legality constraints and an objective combining wirelength, congestion, timing, and power. The textbook wirelength surrogate is the half-perimeter wirelength (HPWL),
\begin{equation}
\label{eq:hpwl}
\mathrm{HPWL}(\mathbf{x},\mathbf{y}) = \sum_{e \in \mathcal{E}}\Big( \max_{v \in e} x_v - \min_{v \in e} x_v + \max_{v \in e} y_v - \min_{v \in e} y_v \Big).
\end{equation}
For differentiable optimization, the non-smooth $\max/\min$ is replaced by a log-sum-exp (LSE) approximation along each axis:
\begin{equation}
\label{eq:lse}
\mathrm{LSE}_x(e;\beta) = \beta \log\!\Big(\sum_{v \in e} e^{x_v / \beta}\Big) + \beta \log\!\Big(\sum_{v \in e} e^{-x_v / \beta}\Big), \qquad
\widetilde{W}(e;\beta) = \mathrm{LSE}_x(e;\beta) + \mathrm{LSE}_y(e;\beta),
\end{equation}
with $\beta>0$ a smoothing parameter; weighted-average wirelength (WA) variants use a different form and we refer to~\cite{lin2019dreamplace,lu2015eplace} for the WA family. Density is enforced via a Poisson penalty $\nabla^2\psi=-\rho$. The full placement objective $\min_{\mathbf{x},\mathbf{y}} \sum_e \widetilde{W}(e;\beta) + \lambda\, \Phi_{\mathrm{dens}}(\mathbf{x},\mathbf{y})$, where $\Phi_{\mathrm{dens}}$ denotes a differentiable density-overflow penalty, is solved by differentiable placers such as ePlace~\cite{lu2015eplace} and DREAMPlace / DREAMPlace 4.0~\cite{lin2019dreamplace,liao2022dreamplace4}; these are \emph{optimization baselines}, not GNN methods, and we list them separately in Table~\ref{tab:eda_apps}. GNN-based macro placement is exemplified by AlphaChip~\cite{mirhoseini2021alphachip}, which combines an edge-based GCN encoder with a reinforcement-learning policy maximizing
\begin{equation}
\label{eq:alphachip_reward}
R = -\big(c_w\, \mathrm{HPWL}_{\text{norm}} + c_c\, \mathrm{Cong}_{\text{norm}} + c_d\, \mathrm{Density}_{\text{norm}}\big),
\end{equation}
with $c_w, c_c, c_d \ge 0$ task-specific weights (distinct from the LSE smoothing $\beta$ above).
AlphaChip remains the most visible public case of graph-based learning for floorplanning, but its evaluation has been debated~\cite{markov2024alphachip_debate,cheng2023assessment} because comparisons depend strongly on benchmark choice, compute budget, human baselines, and proprietary constraints; the original Nature paper carries an editorial note and a 2024 Addendum that names AlphaChip and qualifies parts of the original methodology~\cite{mirhoseini2024addendum}. ChiPFormer~\cite{lai2023chipformer} provides a transformer-based, offline-decision alternative.

\subsection{Routing Congestion}
Global routers discretize the chip into routing-cell (gcell) edges $e_g$ with demand $D(e_g)$ and supply $S(e_g)$:
\begin{equation}
\label{eq:congestion}
\mathrm{Cong}(e_g) = \max\!\big(0,\, D(e_g) - S(e_g)\big), \qquad \mathrm{Overflow} = \sum_{e_g} \mathrm{Cong}(e_g).
\end{equation}
A common analytical demand prior is RUDY (Rectangular Uniform wire DensitY); a simplified form is
\begin{equation}
\label{eq:rudy}
\mathrm{RUDY}(g) = \sum_{e:\, g \in \mathrm{bbox}(e)} \frac{w(e) + h(e)}{w(e)\, h(e)},
\end{equation}
with practical variants normalizing by routing layers, blockages, and net-decomposition rules; degenerate single-row or single-column nets are floored by bin dimensions or an $\epsilon$ term to keep the denominator nonzero. CongestionNet~\cite{kirby2019congestionnet} uses a deep GNN on the netlist graph; LHNN~\cite{wang2022lhnn} respects the lattice-hypergraph structure of nets directly; RouteGNN~\cite{liu2023routegnn} extends to detailed-routing prediction. Resource scarcity at advanced nodes makes early predictions valuable: even post-route fixes can require creative use of middle-of-line metal in filler cells~\cite{kim2021mol_filler}, illustrating how thin design margins have become.

\subsection{Static Timing Analysis}
\label{sec:timing}

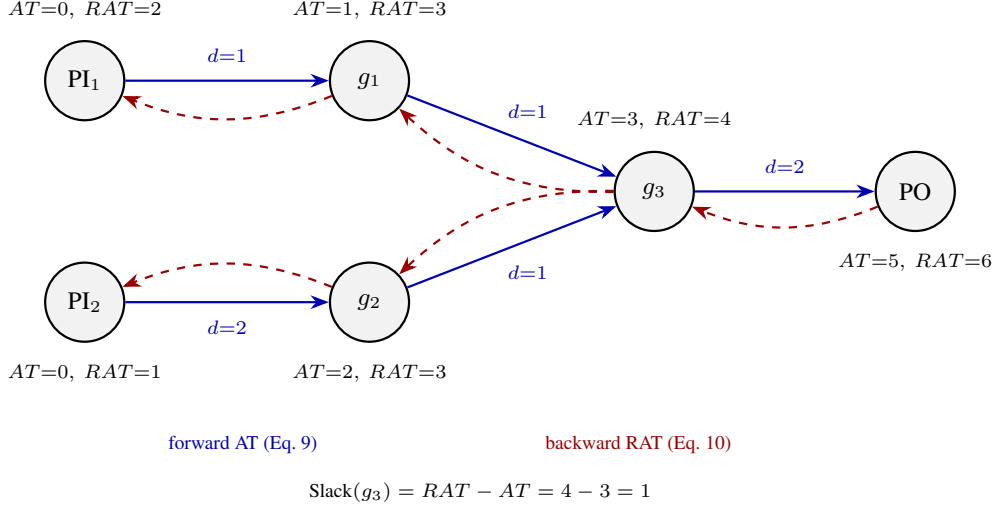
\begin{figure}[t]
\centering
\resizebox{0.95\textwidth}{!}{%
\begin{tikzpicture}[
  every node/.style={font=\small},
  cell/.style={circle,draw,thick,fill=gray!10,minimum size=10mm,inner sep=0pt},
  fwd/.style={->,>=Stealth,thick,blue!70!black},
  bwd/.style={->,>=Stealth,thick,red!60!black,dashed},
  lab/.style={font=\scriptsize},
]
\node[cell] (PI1) at (0,2.0)  {PI$_1$};
\node[cell] (PI2) at (0,-0.8) {PI$_2$};
\node[cell] (G1)  at (3.6,2.0)  {$g_1$};
\node[cell] (G2)  at (3.6,-0.8) {$g_2$};
\node[cell] (G3)  at (7.2,0.6)  {$g_3$};
\node[cell] (PO)  at (10.5,0.6) {PO};
\draw[fwd] (PI1) -- (G1) node[midway,above=1mm,lab] {$d{=}1$};
\draw[fwd] (PI2) -- (G2) node[midway,below=1mm,lab] {$d{=}2$};
\draw[fwd] (G1) -- (G3) node[midway,above=1mm,lab,xshift=2mm] {$d{=}1$};
\draw[fwd] (G2) -- (G3) node[midway,below=1mm,lab,xshift=2mm] {$d{=}1$};
\draw[fwd] (G3) -- (PO) node[midway,above=1mm,lab] {$d{=}2$};
\draw[bwd] (PO) to[bend left=22] (G3);
\draw[bwd] (G3) to[bend left=22] (G1);
\draw[bwd] (G3) to[bend right=22] (G2);
\draw[bwd] (G1) to[bend left=22] (PI1);
\draw[bwd] (G2) to[bend right=22] (PI2);
\node[lab,above=1.5mm of PI1] {$AT{=}0,\ RAT{=}2$};
\node[lab,below=1.5mm of PI2] {$AT{=}0,\ RAT{=}1$};
\node[lab,above=1.5mm of G1]  {$AT{=}1,\ RAT{=}3$};
\node[lab,below=1.5mm of G2]  {$AT{=}2,\ RAT{=}3$};
\node[lab,above=1.5mm of G3]  {$AT{=}3,\ RAT{=}4$};
\node[lab,below=1.5mm of PO]  {$AT{=}5,\ RAT{=}6$};
\node[lab,blue!70!black]  at (2.0,-2.6) {forward AT (Eq.~\ref{eq:at})};
\node[lab,red!60!black]   at (7.0,-2.6) {backward RAT (Eq.~\ref{eq:rat})};
\node[lab] at (5.0,-3.2) {Slack$(g_3)= RAT - AT = 4 - 3 = 1$};
\end{tikzpicture}%
}
\caption{Toy AT/RAT/slack propagation on a combinational DAG with $T_{\text{clk}}=6$. The max-plus / min-plus recurrence structure motivates asynchronous topological message passing (Eq.~\ref{eq:dag_async}); industrial STA additionally handles slew, multi-corner, hold, and OCV derating not shown here.}
\label{fig:atrat}
\end{figure}

The textbook STA recurrences are graph-structured operations on the netlist DAG. Forward-propagated arrival times accumulate worst-case delay; backward-propagated required times accumulate the latest a signal may arrive while still meeting the endpoint requirement $T_{\text{req}}$ (which folds the clock period, endpoint setup time, and clock uncertainty into a single boundary):
\begin{align}
\label{eq:at} AT(v) &= \max_{u \in \mathrm{fanin}(v)}\big(AT(u) + d(u \to v)\big), \quad AT(v_{\text{PI}}) = 0, \\
\label{eq:rat} RAT(v) &= \min_{w \in \mathrm{fanout}(v)}\big(RAT(w) - d(v \to w)\big), \quad RAT(v_{\text{end}}) = T_{\text{req}}, \\
\label{eq:slack} \mathrm{slack}(v) &= RAT(v) - AT(v).
\end{align}
Equation~\ref{eq:at} is a max-plus semiring recurrence; Eq.~\ref{eq:rat} is its min-plus dual. Both motivate asynchronous topological propagation (Eq.~\ref{eq:dag_async}); a synchronous GNN must approximate them with a fixed-depth stack. \emph{This abstraction is single-corner and setup-only}: industrial STA additionally models slew/load-dependent delay through library timing arcs, hold checks, multiple corners, common-path pessimism removal, and on-chip variation derating. Even within this toy abstraction, edge delay decomposes as
\begin{equation}
\label{eq:delay} d(u \to v) = d_{\text{cell}}(u, \mathrm{slew}_u, C_{\text{load}}) + d_{\text{net}}(u \to v),
\end{equation}
motivating two-stage models~\cite{guo2022timing,zhong2024preroutgnn,ye2023timingpredict,ye2023wire_timing}. PreRoutGNN~\cite{zhong2024preroutgnn} predicts pre-routing slack with $R^2 = 0.93$ via global pre-training and attentional cell modeling. Transformer-based delay predictors such as TF-Predictor~\cite{cao2023tfpredictor} and heterogeneous graph-transformer approaches such as ParaFormer~\cite{yoon2025paraformer} predict pre-routing path delay or parasitic RC respectively, feeding into the cell- and net-delay decomposition of Eq.~\ref{eq:delay}; an optimization-aware variant~\cite{he2024opttiming} couples timing prediction directly to downstream optimizer decisions, addressing the actionability gap discussed in \S\ref{sec:open}. Figure~\ref{fig:atrat} visualizes the propagation on a small DAG. A persistent methodological pitfall is \emph{stage leakage}---using post-route or signoff features as inputs to a "pre-route" predictor---which inflates accuracy but offers no real lead time; we return to this in \S\ref{sec:open}.

\subsection{Power and IR Drop}
\label{sec:power}

Per-node dynamic power follows
\begin{equation}
\label{eq:dyn_power}
P_{\text{dyn}}(v) = \alpha_v\, C_v\, V_{dd}^2\, f,
\end{equation}
where $\alpha_v$ is the 0$\to$1 switching activity per cycle. (If $\alpha$ is instead defined as the total transition probability counting both 0$\to$1 and 1$\to$0 events, an equivalent convention introduces a factor of $\frac{1}{2}$.) Estimating $\alpha_v$ at every internal net of a large design is the bottleneck. GRANNITE~\cite{zhang2020grannite} casts switching-activity inference as a learned propagation on the directed netlist,
\begin{equation}
\label{eq:toggle}
\alpha_v^{(\ell+1)} = \phi\!\Big(\alpha_v^{(\ell)},\; \big\{(\alpha_u^{(\ell)}, \tau_u, c_{uv})\big\}_{u \in \mathrm{fanin}(v)}\Big),
\end{equation}
where $\tau_u$ encodes the gate's type or Boolean transfer function and $c_{uv}$ structural correlation context (notation chosen to avoid conflict with the dynamic-power symbol $P_{\mathrm{dyn}}$ above). GRANNITE reports $>18.7\times$ speedup with $<5.5\%$ error against a commercial reference and remains one of the clearest industrial-authored, commercial-reference-evaluated demonstrations of directed GNNs for power estimation. Earlier learning-based power estimators such as PRIMAL~\cite{zhou2019primal} attacked related problems with non-graph-structured ML; we cite them as broader context, not as instances of toggle-rate propagation in the sense above. The power-delivery network admits a complementary linear-system view: $\mathbf{G}\mathbf{v}=\mathbf{i}$ on the conductance Laplacian, with IR drop $\Delta V_{\text{IR}}(n) = V_{dd} - v_n$. PGNN~\cite{pao2023pgnn} embeds the residual $\|\mathbf{G}\mathbf{v}-\mathbf{i}\|$ into the loss; MAVIREC~\cite{chhabria2021mavirec} provides ML-aided vectored estimation.

\subsection{Analog Symmetry Extraction}
\label{sec:analog}

Analog circuits depend on \emph{matching}: differential pairs, current mirrors, and common-centroid arrays must be placed and routed so that systematic mismatch cancels. Identifying these matched groups directly from a schematic netlist is a graph-classification / pairing subproblem for which standard MPNNs have shown useful empirical performance, though downstream layout feasibility still depends on constraint-aware placement and routing. Gao et al.~\cite{gao2021layout} and Chen et al.~\cite{chen2021universal} formulate symmetry / matching extraction as supervised prediction over schematic graphs using standard MPNNs (GCN, GraphSAGE, GAT). The "symmetry" captured here is a discrete pairing constraint between devices, not a continuous rotation/reflection---the relevant group is a finite subgroup of $E(2)$, and standard message passing plus augmentation over the small legal orientation set is often a reasonable modeling choice for this subproblem. ALIGN~\cite{kunal2020align} is an open-source analog-layout flow that consumes such constraints; it is a flow framework rather than a GNN method, and we list it as such in Table~\ref{tab:eda_apps}. CktGNN~\cite{dong2023cktgnn} formulates analog topology search itself as a graph-generation problem, and recent analog-topology efforts increasingly combine GNN encoders with diffusion or LLM decoders.

\begin{table}[t]
\centering
\caption{Representative methods across EDA tasks and flow contexts. The evidence column intentionally separates academic-benchmark research, non-GNN baselines, flow frameworks, industrial-authored evaluations, and publicly reported production use, so that GNN-specific deployment evidence is not conflated with broader ML-EDA adoption.}
\label{tab:eda_apps}
\footnotesize
\setlength{\tabcolsep}{4pt}
\resizebox{\textwidth}{!}{%
\begin{tabular}{@{}lllll@{}}
\toprule
\textbf{Stage} & \textbf{Method} & \textbf{Type} & \textbf{Learning target} & \textbf{Evidence} \\
\midrule
Logic synthesis & DeepGate / DeepGate2 / DeepGate3~\cite{li2022deepgate,shi2023deepgate2,shi2024deepgate3} & GNN on AIG & Functional + structural embed. & Research \\
Logic synthesis & Gamora~\cite{wu2023gamora} & Multi-task GNN & Operator recovery & Research \\
Logic synthesis & PolarGate~\cite{liu2024polargate} & Polarity-aware DAG-GNN & AIG functional embedding & Research \\
Floorplanning & AlphaChip~\cite{mirhoseini2021alphachip,mirhoseini2024addendum} & GCN + RL policy & Macro placement & Publicly reported production use* \\
Floorplanning & ChiPFormer~\cite{lai2023chipformer} & Decision Transformer & Cross-design transfer & Research \\
Placement (baseline) & ePlace / DREAMPlace / DREAMPlace 4.0~\cite{lu2015eplace,lin2019dreamplace,liao2022dreamplace4} & Differentiable placer & Wirelength + density opt. & Non-GNN baseline \\
Congestion & CongestionNet / LHNN / RouteGNN~\cite{kirby2019congestionnet,wang2022lhnn,liu2023routegnn} & GNN / hyper-GNN & Demand / overflow prediction & Research \\
Timing & PreRoutGNN, Guo et al.~\cite{zhong2024preroutgnn,guo2022timing} & Async DAG-GNN & Pre-route slack & Research \\
Timing & TimingPredict, ParaFormer~\cite{ye2023timingpredict,ye2023wire_timing,yoon2025paraformer} & GNN / hetero-GT & Path / wire delay, parasitic RC & Research \\
Power & GRANNITE~\cite{zhang2020grannite} & Directed GNN & Switching activity & Industrial-authored, public evaluation \\
IR drop & PGNN, MAVIREC~\cite{pao2023pgnn,chhabria2021mavirec} & Physics-informed / ML & Voltage drop, hotspot & Research \\
MBFF clustering & Jeong \& Kim~\cite{jeong2025mlcad_mbff} & GNN on FF graph & Power-aware clustering & Research \\
Analog symmetry & Gao et al., Chen et al.~\cite{gao2021layout,chen2021universal} & GNN on schematic & Constraint extraction & Research \\
Analog flow (context) & ALIGN~\cite{kunal2020align} & Open-source flow & SPICE-to-GDS automation & Flow framework \\
Multi-die (context) & Pin-3D~\cite{pentapati2023pin3d} & Physical-design flow & 3D physical design & Flow context \\
\bottomrule
\end{tabular}%
}
\\[1mm]
{\scriptsize *Public industrial reports are not equivalent to open, independently reproducible benchmarks. ``Industrial-authored, public evaluation'' indicates that an industrial group has authored the method and reported results against a commercial-flow reference, without an explicit production-deployment claim.}
\end{table}

\section{Industrial Evidence and the Benchmark Gap}
\label{sec:industrial}

It is tempting to read the union of GNN-for-EDA results plus the integration of machine learning into commercial flows as evidence of broad industrial GNN adoption. The two should be separated. \emph{GNN-specific} public evidence is narrow and uneven: AlphaChip~\cite{mirhoseini2021alphachip,mirhoseini2024addendum} is a publicly reported production floorplanning case, while GRANNITE~\cite{zhang2020grannite} is an industrial-authored power-estimation result evaluated against commercial-flow references with reported speed and accuracy gains, but not accompanied by a comparably explicit production-deployment claim. The same task-alignment lens explains why the strongest public results---both industrial cases and high-performing task-specific research---are stage-specific rather than universal: floorplanning, power estimation, and timing prediction each expose different algebraic structure and different available features, and a single backbone for all of them does not yet exist. \emph{Broader ML-EDA} evidence is wider---Synopsys' DSO.ai and Cadence's Cerebrus are well-publicized commercial offerings---but their internal model classes are not publicly disclosed, and reading them as endorsements of GNNs in particular is unwarranted. Earlier broad surveys of ML-for-EDA~\cite{huang2021mleda} and book-length treatments~\cite{ren2023mleda_book} document this distinction.

\paragraph{Benchmarks.} Reproducibility has been a persistent bottleneck. Most industrial netlists are proprietary, and academic GNN-for-EDA results have historically been validated on small open designs whose statistics differ markedly from production netlists. CircuitNet~\cite{chai2022circuitnet} provides an open benchmark with thousands of RISC-V samples for congestion, IR-drop, and timing-related prediction; CircuitNet 2.0~\cite{jiang2024circuitnet2} extends to 28\,nm and 14\,nm CPU, GPU, and AI-chip blocks, but task availability differs across subsets (for instance, the 14\,nm subset supports congestion, IR drop, and net delay, but DRC violation prediction is not currently supported). The empirical gap between dense-prediction tasks on current benchmarks and rare-event signoff tasks such as IR-drop hotspot recall or DRC violation prediction is precisely what motivates the proxy-to-signoff and calibration challenges in \S\ref{sec:open}; we view CircuitNet's continued development as a high-leverage activity, particularly given the access asymmetry created by largely proprietary industrial flows and signoff data.

\section{Open Challenges}
\label{sec:open}

We highlight the issues we believe are likely to dominate the next phase of GNN-for-EDA work. Each is more concrete than "more architectures" and is tied to the algebra--task framework of \S\ref{sec:applications}.

\paragraph{Stage leakage.} A pre-routing prediction is only useful if its inputs are actually available before routing. Many published pre-route timing or congestion models are trained or evaluated using post-route features (parasitics, post-route congestion maps), inflating reported accuracy without offering real lead time. Auditing predictors at the stage where their inputs are obtainable is a basic but underemphasized hygiene requirement.

\paragraph{Proxy-to-signoff gap.} Models trained against proxy targets---HPWL, RUDY, early-stage slack estimates---can over-fit to those proxies and underperform on signoff QoR. Joint training against signoff metrics, or calibration of proxy-trained predictors against signoff distributions, is largely open.

\paragraph{Calibration and uncertainty.} In EDA, a confidently wrong prediction at signoff is far more costly than an uncertain hedge. GNNs in EDA are routinely reported with pointwise MAE / $R^2$ but rarely with calibrated uncertainty intervals or rare-event metrics (timing-violation tail, EM/IR hotspot recall). Bayesian or ensembling approaches, rare-event-aware loss design, and conformal prediction are under-explored; Bayesian GNN posteriors such as the Graph Posterior Network~\cite{stadler2021gpn} are a natural starting point. Concretely, future EDA-GNN benchmarking should report 95\% confidence intervals over multiple seeds, paired non-parametric tests (e.g., Wilcoxon signed-rank) against baselines, and rare-event metrics (tail recall, hotspot precision-at-$K$) alongside pointwise MAE/$R^2$.

\paragraph{Distribution shift across PDK / library / tool.} A model trained on 28\,nm designs typically does not generalize to 5\,nm without retraining, and the labeled data needed for retraining is precisely what is scarce. Cross-design, cross-PDK, cross-library, and cross-tool generalization---and how much of a model is genuinely transferable versus must be re-fitted---is the central open question for industrial deployment. Causal-invariance and OOD methods for graphs (e.g., CIGA~\cite{chen2022ciga}), graph structure-learning approaches against structural noise (Pro-GNN~\cite{jin2020prognn}), and topology-imbalance mitigations (GraphSMOTE~\cite{zhao2021graphsmote}, ReNode~\cite{chen2021renode}) developed for citation networks address recognizable EDA problems and are under-exploited here.

\paragraph{Scale: a systems issue, not only a modeling one.} Beyond model scaling, training heterogeneous circuit GNNs on industrial designs is itself a systems bottleneck. DR-CircuitGNN~\cite{luo2025drcircuitgnn} reports that GPU-kernel-level optimization of heterogeneous-graph forward and backward passes is necessary to make CircuitNet-scale training tractable, suggesting that EDA-GNN scalability progress will require attention to both algorithmic abstraction and infrastructure. Industrial-scale graph engineering at the level of PinSage~\cite{ying2018pinsage} (3\,B nodes, 17\,B edges) is instructive about what such infrastructure looks like in another domain.

\paragraph{Actionability.} A predictor that only produces a number is less useful than one that exposes the gradient or ranking that an optimizer can act on. Aligning predictor outputs with the inputs that downstream optimizers (placers, routers, sizers) consume is a software-architecture problem as much as a learning problem.

\paragraph{Multi-die regime.} 2.5D and 3D integration introduce graph hierarchies that current single-die models do not address. Heterogeneous GNNs at chiplet$\to$die$\to$cell scale---and benchmarks for the same---are nearly absent.

\paragraph{Foundation models for circuits.} Public evidence so far does not establish that cross-domain graph foundation models~\cite{liu2024ofa,xia2024anygraph} match specialist EDA models on stage-specific tasks; the heterogeneity between AIGs, heterogeneous netlists, and PDN graphs may resist any single representation. Recent circuit-specific encoders (CircuitFusion~\cite{fang2025circuitfusion}, NetTAG~\cite{fang2025nettag}, stage-aligned encoders~\cite{fang2025circuitencoder}) move beyond AIG-only models, suggesting the most promising near-term direction is narrower than ``circuit GPT'': stage-specific or modality-aligned backbones rather than unified ones. Domain-adapted LLMs for chip design~\cite{liu2024chipnemo} are an orthogonal but related line.

\section{Conclusion}

The strongest public GNN-for-EDA results arise when graph computation respects the algebra of the target task: max-plus/min-plus DAGs for timing, hypergraph wirelength for placement, demand-supply fields for congestion, probabilistic recurrences for switching activity, Laplacian systems for IR drop, and discrete constraints for analog symmetry. The open agenda---stage leakage, proxy-to-signoff gap, calibration, distribution shift, scale, multi-die---reflects not a crisis but maturity.


\bibliographystyle{unsrt}

\begin{thebibliography}{99}

\bibitem{wu2021comprehensive}
Z.~Wu, S.~Pan, F.~Chen, G.~Long, C.~Zhang, and P.~S. Yu.
\newblock A comprehensive survey on graph neural networks.
\newblock \emph{IEEE Trans. Neural Netw. Learn. Syst.}, 32(1):4--24, 2021.

\bibitem{ju2025realworld}
W.~Ju, S.~Yi, Y.~Wang, Z.~Xiao, Z.~Mao, H.~Li, Y.~Gu, Y.~Qin, N.~Yin, S.~Wang, X.~Liu, P.~S. Yu, and M.~Zhang.
\newblock A survey of graph neural networks in real world: Imbalance, noise, privacy and {OOD} challenges.
\newblock \emph{arXiv:2403.04468v2}, 2025.

\bibitem{huang2021mleda}
G.~Huang, J.~Hu, Y.~He, J.~Liu, M.~Ma, Z.~Shen, J.~Wu, Y.~Xu, H.~Zhang, K.~Zhong, X.~Ning, Y.~Ma, H.~Yang, B.~Yu, H.~Yang, and Y.~Wang.
\newblock Machine learning for electronic design automation: A survey.
\newblock \emph{ACM TODAES}, 26(5):40:1--40:46, 2021.

\bibitem{ren2023mleda_book}
H.~Ren and J.~Hu (Eds.).
\newblock \emph{Machine Learning Applications in Electronic Design Automation}.
\newblock Springer, 2023.

\bibitem{bruna2014}
J.~Bruna, W.~Zaremba, A.~Szlam, and Y.~LeCun.
\newblock Spectral networks and locally connected networks on graphs.
\newblock In \emph{Proc. ICLR}, 2014.

\bibitem{defferrard2016}
M.~Defferrard, X.~Bresson, and P.~Vandergheynst.
\newblock Convolutional neural networks on graphs with fast localized spectral filtering.
\newblock In \emph{Proc. NeurIPS}, pp.~3837--3845, 2016.

\bibitem{kipf2017}
T.~N. Kipf and M.~Welling.
\newblock Semi-supervised classification with graph convolutional networks.
\newblock In \emph{Proc. ICLR}, 2017.

\bibitem{gilmer2017}
J.~Gilmer, S.~S. Schoenholz, P.~F. Riley, O.~Vinyals, and G.~E. Dahl.
\newblock Neural message passing for quantum chemistry.
\newblock In \emph{Proc. ICML}, pp.~1263--1272, 2017.

\bibitem{hamilton2017}
W.~L. Hamilton, Z.~Ying, and J.~Leskovec.
\newblock Inductive representation learning on large graphs.
\newblock In \emph{Proc. NeurIPS}, 2017.

\bibitem{velickovic2018}
P.~Veli\v{c}kovi\'{c}, G.~Cucurull, A.~Casanova, A.~Romero, P.~Li\`{o}, and Y.~Bengio.
\newblock Graph attention networks.
\newblock In \emph{Proc. ICLR}, 2018.

\bibitem{brody2022gatv2}
S.~Brody, U.~Alon, and E.~Yahav.
\newblock How attentive are graph attention networks?
\newblock In \emph{Proc. ICLR}, 2022.

\bibitem{xu2019gin}
K.~Xu, W.~Hu, J.~Leskovec, and S.~Jegelka.
\newblock How powerful are graph neural networks?
\newblock In \emph{Proc. ICLR}, 2019.

\bibitem{schlichtkrull2018rgcn}
M.~Schlichtkrull, T.~N. Kipf, P.~Bloem, R.~van~den~Berg, I.~Titov, and M.~Welling.
\newblock Modeling relational data with graph convolutional networks.
\newblock In \emph{Proc. ESWC}, pp.~593--607, 2018.

\bibitem{hu2020hgt}
Z.~Hu, Y.~Dong, K.~Wang, and Y.~Sun.
\newblock Heterogeneous graph transformer.
\newblock In \emph{Proc. WWW}, pp.~2704--2710, 2020.

\bibitem{feng2019hgnn}
Y.~Feng, H.~You, Z.~Zhang, R.~Ji, and Y.~Gao.
\newblock Hypergraph neural networks.
\newblock In \emph{Proc. AAAI}, 33(01):3558--3565, 2019.

\bibitem{ying2018diffpool}
R.~Ying, J.~You, C.~Morris, X.~Ren, W.~L. Hamilton, and J.~Leskovec.
\newblock Hierarchical graph representation learning with differentiable pooling.
\newblock In \emph{Proc. NeurIPS}, pp.~4800--4810, 2018.

\bibitem{lee2019sagpool}
J.~Lee, I.~Lee, and J.~Kang.
\newblock Self-attention graph pooling.
\newblock In \emph{Proc. ICML}, pp.~3734--3743, 2019.

\bibitem{morris2019kwl}
C.~Morris, M.~Ritzert, M.~Fey, W.~L. Hamilton, J.~E. Lenssen, G.~Rattan, and M.~Grohe.
\newblock Weisfeiler and {Leman} go neural: Higher-order graph neural networks.
\newblock In \emph{Proc. AAAI}, 33(01):4602--4609, 2019.

\bibitem{maron2019ppgn}
H.~Maron, H.~Ben-Hamu, H.~Serviansky, and Y.~Lipman.
\newblock Provably powerful graph networks.
\newblock In \emph{Proc. NeurIPS}, pp.~2153--2164, 2019.

\bibitem{li2018oversmoothing}
Q.~Li, Z.~Han, and X.-M. Wu.
\newblock Deeper insights into graph convolutional networks for semi-supervised learning.
\newblock In \emph{Proc. AAAI}, pp.~3538--3545, 2018.

\bibitem{oono2020oversmoothing}
K.~Oono and T.~Suzuki.
\newblock Graph neural networks exponentially lose expressive power for node classification.
\newblock In \emph{Proc. ICLR}, 2020.

\bibitem{xu2018jknet}
K.~Xu, C.~Li, Y.~Tian, T.~Sonobe, K.~Kawarabayashi, and S.~Jegelka.
\newblock Representation learning on graphs with jumping knowledge networks.
\newblock In \emph{Proc. ICML}, 2018.

\bibitem{chen2020gcnii}
M.~Chen, Z.~Wei, Z.~Huang, B.~Ding, and Y.~Li.
\newblock Simple and deep graph convolutional networks.
\newblock In \emph{Proc. ICML}, pp.~1725--1735, 2020.

\bibitem{zhao2020pairnorm}
L.~Zhao and L.~Akoglu.
\newblock {PairNorm}: Tackling oversmoothing in {GNN}s.
\newblock In \emph{Proc. ICLR}, 2020.

\bibitem{rong2020dropedge}
Y.~Rong, W.~Huang, T.~Xu, and J.~Huang.
\newblock {DropEdge}: Towards deep graph convolutional networks on node classification.
\newblock In \emph{Proc. ICLR}, 2020.

\bibitem{alon2021oversquashing}
U.~Alon and E.~Yahav.
\newblock On the bottleneck of graph neural networks and its practical implications.
\newblock In \emph{Proc. ICLR}, 2021.

\bibitem{topping2022oversquashing}
J.~Topping, F.~Di~Giovanni, B.~P. Chamberlain, X.~Dong, and M.~M. Bronstein.
\newblock Understanding over-squashing and bottlenecks on graphs via curvature.
\newblock In \emph{Proc. ICLR}, 2022.

\bibitem{zhang2019dvae}
M.~Zhang, S.~Jiang, Z.~Cui, R.~Garnett, and Y.~Chen.
\newblock {D-VAE}: A variational autoencoder for directed acyclic graphs.
\newblock In \emph{Proc. NeurIPS}, 2019.

\bibitem{thost2021dagnn}
V.~Thost and J.~Chen.
\newblock Directed acyclic graph neural networks.
\newblock In \emph{Proc. ICLR}, 2021.

\bibitem{ying2021graphormer}
C.~Ying, T.~Cai, S.~Luo, S.~Zheng, G.~Ke, D.~He, Y.~Shen, and T.-Y. Liu.
\newblock Do transformers really perform badly for graph representation?
\newblock In \emph{Proc. NeurIPS}, 2021.

\bibitem{rampasek2022gps}
L.~Ramp\'{a}\v{s}ek, M.~Galkin, V.~P. Dwivedi, A.~T. Luu, G.~Wolf, and D.~Beaini.
\newblock Recipe for a general, powerful, scalable graph transformer.
\newblock In \emph{Proc. NeurIPS}, 2022.

\bibitem{wu2022nodeformer}
Q.~Wu, W.~Zhao, Z.~Li, D.~P. Wipf, and J.~Yan.
\newblock {NodeFormer}: A scalable graph structure learning transformer for node classification.
\newblock In \emph{Proc. NeurIPS}, 2022.

\bibitem{shirzad2023exphormer}
H.~Shirzad, A.~Velingker, B.~Venkatachalam, D.~J. Sutherland, and A.~K. Sinop.
\newblock Exphormer: Sparse transformers for graphs.
\newblock In \emph{Proc. ICML}, 2023.

\bibitem{wu2023sgformer}
Q.~Wu, W.~Zhao, C.~Yang, H.~Zhang, F.~Nie, H.~Jiang, Y.~Bian, and J.~Yan.
\newblock {SGFormer}: Simplifying and empowering transformers for large-graph representations.
\newblock In \emph{Proc. NeurIPS}, 2023.

\bibitem{deng2024polynormer}
C.~Deng, Y.~Li, Z.~Feng, H.~Ren, and Z.~Zhang.
\newblock Polynormer: Polynomial-expressive graph transformer in linear time.
\newblock In \emph{Proc. ICLR}, 2024.

\bibitem{schutt2017schnet}
K.~T. Sch\"{u}tt, P.-J. Kindermans, H.~E. Sauceda, S.~Chmiela, A.~Tkatchenko, and K.-R. M\"{u}ller.
\newblock {SchNet}: A continuous-filter convolutional neural network for modeling quantum interactions.
\newblock In \emph{Proc. NeurIPS}, pp.~991--1001, 2017.

\bibitem{satorras2021egnn}
V.~G. Satorras, E.~Hoogeboom, and M.~Welling.
\newblock {E(n)} equivariant graph neural networks.
\newblock In \emph{Proc. ICML}, pp.~9323--9332, 2021.

\bibitem{batatia2022mace}
I.~Batatia, D.~P. Kov\'{a}cs, G.~N.~C. Simm, C.~Ortner, and G.~Cs\'{a}nyi.
\newblock {MACE}: Higher order equivariant message passing neural networks for fast and accurate force fields.
\newblock In \emph{Proc. NeurIPS}, pp.~11423--11436, 2022.

\bibitem{liao2024equiformerv2}
Y.-L. Liao, B.~Wood, A.~Das, and T.~Smidt.
\newblock {EquiformerV2}: Improved equivariant transformer for scaling to higher-degree representations.
\newblock In \emph{Proc. ICLR}, 2024.

\bibitem{li2022deepgate}
M.~Li, S.~Khan, Z.~Shi, N.~Wang, H.~Yu, and Q.~Xu.
\newblock {DeepGate}: Learning neural representations of logic gates.
\newblock In \emph{Proc. DAC}, 2022.

\bibitem{shi2023deepgate2}
Z.~Shi, H.~Pan, S.~Khan, M.~Li, Y.~Liu, J.~Huang, H.-L. Zhen, M.~Yuan, Z.~Chu, and Q.~Xu.
\newblock {DeepGate2}: Functionality-aware circuit representation learning.
\newblock In \emph{Proc. ICCAD}, 2023.

\bibitem{shi2024deepgate3}
Z.~Shi, Z.~Zheng, S.~Khan, J.~Zhong, M.~Li, and Q.~Xu.
\newblock {DeepGate3}: Towards scalable circuit representation learning.
\newblock In \emph{Proc. ICCAD}, 2024.

\bibitem{wu2023gamora}
N.~Wu, Y.~Li, C.~Hao, S.~Dai, C.~Yu, and Y.~Xie.
\newblock Gamora: Graph learning based symbolic reasoning for large-scale {Boolean} networks.
\newblock In \emph{Proc. DAC}, 2023.

\bibitem{liu2024polargate}
J.~Liu, J.~Zhai, M.~Zhao, Z.~Lin, B.~Yu, and C.~Shi.
\newblock {PolarGate}: Breaking the functionality representation bottleneck of {And}-{Inverter} graph neural network.
\newblock In \emph{Proc. ICCAD}, 2024.

\bibitem{lin2019dreamplace}
Y.~Lin, S.~Dhar, W.~Li, H.~Ren, B.~Khailany, and D.~Z. Pan.
\newblock {DREAMPlace}: Deep learning toolkit-enabled {GPU} acceleration for modern {VLSI} placement.
\newblock In \emph{Proc. DAC}, 2019.

\bibitem{liao2022dreamplace4}
P.~Liao, S.~Liu, Z.~Chen, W.~Lv, Y.~Lin, and B.~Yu.
\newblock {DREAMPlace 4.0}: Timing-driven global placement with momentum-based net weighting.
\newblock In \emph{Proc. DATE}, 2022.

\bibitem{lu2015eplace}
J.~Lu, P.~Chen, C.-C. Chang, L.~Sha, D.~J.-H. Huang, C.-C. Teng, and C.-K. Cheng.
\newblock {ePlace}: Electrostatics-based placement using fast {Fourier} transform and {Nesterov}'s method.
\newblock \emph{ACM TODAES}, 20(2):17:1--17:34, 2015.

\bibitem{mirhoseini2021alphachip}
A.~Mirhoseini, A.~Goldie, M.~Yazgan, J.~W. Jiang, E.~M. Songhori, S.~Wang, Y.-J. Lee, E.~Johnson, O.~Pathak, A.~Nova, J.~Pak, A.~Tong, K.~Srinivasa, W.~Hang, E.~Tuncer, Q.~V. Le, J.~Laudon, R.~Ho, R.~Carpenter, and J.~Dean.
\newblock A graph placement methodology for fast chip design.
\newblock \emph{Nature}, 594:207--212, 2021.

\bibitem{mirhoseini2024addendum}
A.~Goldie, A.~Mirhoseini, M.~Yazgan, et~al.
\newblock Addendum: A graph placement methodology for fast chip design.
\newblock \emph{Nature}, 634:E10--E11, 2024.

\bibitem{markov2024alphachip_debate}
I.~L. Markov, D.~Lee, and C.-K. Cheng.
\newblock Is chip floorplanning a solved problem?
\newblock \emph{Commun. ACM}, 67(11):66--74, 2024.

\bibitem{cheng2023assessment}
C.-K. Cheng, A.~B. Kahng, S.~Kundu, Y.~Wang, and Z.~Wang.
\newblock Assessment of reinforcement learning for macro placement.
\newblock In \emph{Proc. ISPD}, 2023.

\bibitem{lai2023chipformer}
Y.~Lai, J.~Liu, Z.~Tang, B.~Wang, J.~Hao, and P.~Luo.
\newblock {ChiPFormer}: Transferable chip placement via offline decision transformer.
\newblock In \emph{Proc. ICML}, 2023.

\bibitem{kirby2019congestionnet}
R.~Kirby, S.~Godil, R.~Roy, and B.~Catanzaro.
\newblock {CongestionNet}: Routing congestion prediction using deep graph neural networks.
\newblock In \emph{Proc. IFIP/IEEE International Conference on Very Large Scale Integration (VLSI-SoC)}, 2019.

\bibitem{kim2021mol_filler}
J.~Jeong and T.~Kim.
\newblock Utilizing middle-of-line resource in filler cells for fixing routing failures.
\newblock In \emph{Proc. IEEE/ACM International Conference on Computer-Aided Design (ICCAD)}, 2021.

\bibitem{wang2022lhnn}
B.~Wang, G.~Shen, D.~Li, J.~Hao, W.~Liu, Y.~Huang, H.~Wu, Y.~Lin, G.~Chen, and P.~A. Heng.
\newblock {LHNN}: Lattice hypergraph neural network for {VLSI} congestion prediction.
\newblock In \emph{Proc. DAC}, 2022.

\bibitem{liu2023routegnn}
S.~Liu, P.~Liao, P.~Sun, Y.~Lin, and B.~Yu.
\newblock {RouteGNN}: Learning to route via graph neural networks.
\newblock In \emph{Proc. ICCAD}, 2023.

\bibitem{zhong2024preroutgnn}
R.~Zhong, J.~Ye, Z.~Tang, S.~Kai, M.~Yuan, J.~Hao, and J.~Yan.
\newblock {PreRoutGNN} for timing prediction with order preserving partition.
\newblock In \emph{Proc. AAAI}, 38(15):17087--17095, 2024.

\bibitem{guo2022timing}
Z.~Guo, M.~Liu, J.~Gu, S.~Zhang, D.~Z. Pan, and Y.~Lin.
\newblock A timing engine inspired graph neural network model for pre-routing slack prediction.
\newblock In \emph{Proc. DAC}, 2022.

\bibitem{ye2023timingpredict}
Y.~Ye, T.~Chen, Y.~Gao, H.~Yan, B.~Yu, and L.~Shi.
\newblock {TimingPredict}: {GNN}-based pre-routing path delay prediction.
\newblock In \emph{Proc. ICCAD}, 2023.

\bibitem{ye2023wire_timing}
Y.~Ye, T.~Chen, Y.~Gao, H.~Yan, B.~Yu, and L.~Shi.
\newblock Fast and accurate wire timing estimation on tree and non-tree net structures.
\newblock In \emph{Proc. DAC}, 2023.

\bibitem{cao2023tfpredictor}
P.~Cao, G.~He, and T.~Yang.
\newblock {TF-Predictor}: Transformer-based prerouting path delay prediction framework.
\newblock \emph{IEEE Trans. Comput.-Aided Design Integr. Circuits Syst.}, 42(7):2227--2237, 2023.

\bibitem{zhang2020grannite}
Y.~Zhang, H.~Ren, and B.~Khailany.
\newblock {GRANNITE}: Graph neural network inference for transferable power estimation.
\newblock In \emph{Proc. DAC}, pp.~1--6, 2020.

\bibitem{zhou2019primal}
Y.~Zhou, H.~Ren, Y.~Zhang, B.~Keller, B.~Khailany, and Z.~Zhang.
\newblock {PRIMAL}: Power inference using machine learning.
\newblock In \emph{Proc. DAC}, 2019.

\bibitem{pao2023pgnn}
Y.~C. Pao and M.~D.~F. Wong.
\newblock {PGNN}: Physics-inspired graph neural network for dynamic {IR}-drop prediction.
\newblock In \emph{Proc. DAC}, 2023.

\bibitem{chhabria2021mavirec}
V.~A. Chhabria, V.~Ahuja, A.~Prabhu, N.~Patil, P.~Jain, and S.~S. Sapatnekar.
\newblock {MAVIREC}: {ML}-aided vectored {IR}-drop estimation and classification.
\newblock In \emph{Proc. DATE}, 2021.

\bibitem{jeong2025mlcad_mbff}
J.~Jeong and T.~Kim.
\newblock Machine learning driven early clustering for multi-bit flip-flop allocation.
\newblock In \emph{Proc. ACM/IEEE Symposium on Machine Learning for CAD (MLCAD)}, 2025.

\bibitem{jeong2024placement_legalization}
J.~Jeong and T.~Kim.
\newblock Placement legalization for heterogeneous cells of non-integer multiple-heights.
\newblock \emph{Integration}, 97:102177, 2024.

\bibitem{jeong2024dtco_mbff}
J.~Jeong and T.~Kim.
\newblock Binding multi-bit flip-flop cells through design and technology co-optimization.
\newblock In \emph{Proc. DAC}, 2024.

\bibitem{kunal2020align}
K.~Kunal, M.~Madhusudan, A.~K. Sharma, W.~Xu, S.~M. Burns, R.~Harjani, J.~Hu, D.~A. Kirkpatrick, and S.~S. Sapatnekar.
\newblock {ALIGN}: Open-source analog layout automation from the ground up.
\newblock In \emph{Proc. DAC}, 2020.

\bibitem{gao2021layout}
X.~Gao, C.~Deng, M.~Liu, Z.~Zhang, D.~Z. Pan, and Y.~Lin.
\newblock Layout symmetry annotation for analog circuits with graph neural networks.
\newblock In \emph{Proc. ASP-DAC}, 2021.

\bibitem{chen2021universal}
H.~Chen, K.~Zhu, M.~Liu, X.~Tang, N.~Sun, and D.~Z. Pan.
\newblock Universal symmetry constraint extraction for analog and mixed-signal circuits with graph neural networks.
\newblock In \emph{Proc. DAC}, 2021.

\bibitem{dong2023cktgnn}
Z.~Dong, W.~Cao, M.~Zhang, D.~Tao, Y.~Chen, and X.~Zhang.
\newblock {CktGNN}: Circuit graph neural network for electronic design automation.
\newblock In \emph{Proc. ICLR}, 2023.

\bibitem{pentapati2023pin3d}
S.~M.~P.~D. Pentapati, K.~Chang, V.~Gerousis, R.~Sengupta, and S.~K. Lim.
\newblock {Pin-3D}: A physical synthesis and post-layout optimization flow for heterogeneous 3D {ICs}.
\newblock In \emph{Proc. ICCAD}, 2023.

\bibitem{chai2022circuitnet}
Z.~Chai, Y.~Zhao, Y.~Lin, W.~Liu, R.~Wang, and R.~Huang.
\newblock {CircuitNet}: An open-source dataset for machine learning applications in electronic design automation.
\newblock \emph{Sci. China Inf. Sci.}, 65(12):227401, 2022.

\bibitem{jiang2024circuitnet2}
X.~Jiang, Z.~Chai, Y.~Zhao, Y.~Lin, R.~Wang, and R.~Huang.
\newblock {CircuitNet 2.0}: An advanced dataset for promoting machine learning in {VLSI} {EDA}.
\newblock In \emph{Proc. ICLR}, 2024.

\bibitem{liu2024chipnemo}
M.~Liu, T.-D. Ene, R.~Kirby, C.~Cheng, N.~Pinckney, R.~Liang, J.~Alben, H.~Anand, S.~Banerjee, et~al.
\newblock {ChipNeMo}: Domain-adapted {LLMs} for chip design.
\newblock arXiv:2311.00176, 2024.

\bibitem{liu2024ofa}
H.~Liu, J.~Feng, L.~Kong, N.~Liang, D.~Tao, Y.~Chen, and M.~Zhang.
\newblock One for all: Towards training one graph model for all classification tasks.
\newblock In \emph{Proc. ICLR}, 2024.

\bibitem{xia2024anygraph}
L.~Xia and C.~Huang.
\newblock {AnyGraph}: Graph foundation model in the wild.
\newblock arXiv:2408.10700, 2024.

\bibitem{gasse2019milp}
M.~Gasse, D.~Ch\'{e}telat, N.~Ferroni, L.~Charlin, and A.~Lodi.
\newblock Exact combinatorial optimization with graph convolutional neural networks.
\newblock In \emph{Proc. NeurIPS}, 2019.

\bibitem{ying2018pinsage}
R.~Ying, R.~He, K.~Chen, P.~Eksombatchai, W.~L. Hamilton, and J.~Leskovec.
\newblock Graph convolutional neural networks for web-scale recommender systems.
\newblock In \emph{Proc. KDD}, pp.~974--983, 2018.

\bibitem{zhao2021graphsmote}
T.~Zhao, X.~Zhang, and S.~Wang.
\newblock {GraphSMOTE}: Imbalanced node classification on graphs with graph neural networks.
\newblock In \emph{Proc. WSDM}, pp.~833--841, 2021.

\bibitem{chen2021renode}
D.~Chen, Y.~Lin, G.~Zhao, X.~Ren, P.~Li, J.~Zhou, and X.~Sun.
\newblock Topology-imbalance learning for semi-supervised node classification.
\newblock In \emph{Proc. NeurIPS}, 34:29885--29897, 2021.

\bibitem{jin2020prognn}
W.~Jin, Y.~Ma, X.~Liu, X.~Tang, S.~Wang, and J.~Tang.
\newblock Graph structure learning for robust graph neural networks.
\newblock In \emph{Proc. KDD}, pp.~66--74, 2020.

\bibitem{chen2022ciga}
Y.~Chen, Y.~Zhang, Y.~Bian, H.~Yang, K.~Ma, B.~Xie, T.~Liu, B.~Han, and J.~Cheng.
\newblock Learning causally invariant representations for out-of-distribution generalization on graphs.
\newblock In \emph{Proc. NeurIPS}, 2022.

\bibitem{stadler2021gpn}
M.~Stadler, B.~Charpentier, S.~Geisler, D.~Z\"{u}gner, and S.~G\"{u}nnemann.
\newblock Graph posterior network: Bayesian predictive uncertainty for node classification.
\newblock In \emph{Proc. NeurIPS}, pp.~18033--18048, 2021.

\bibitem{fang2025circuitfusion}
W.~Fang, S.~Liu, J.~Wang, and Z.~Xie.
\newblock {CircuitFusion}: Multimodal circuit representation learning for agile chip design.
\newblock In \emph{Proc. ICLR}, 2025.

\bibitem{wu2025mgvga}
H.~Wu, H.~Zheng, Y.~Pu, and B.~Yu.
\newblock Circuit representation learning with masked gate modeling and {Verilog}-{AIG} alignment.
\newblock In \emph{Proc. ICLR}, 2025.

\bibitem{fang2025nettag}
W.~Fang, W.~Li, S.~Liu, Y.~Lu, H.~Zhang, and Z.~Xie.
\newblock {NetTAG}: A multimodal {RTL}-and-layout-aligned netlist foundation model via text-attributed graph.
\newblock In \emph{Proc. DAC}, 2025.
\newblock DOI: \url{10.1109/DAC63849.2025.11133349}.

\bibitem{fang2025circuitencoder}
W.~Fang, S.~Liu, H.~Zhang, and Z.~Xie.
\newblock A self-supervised, pre-trained, and cross-stage-aligned circuit encoder provides a foundation for various design tasks.
\newblock In \emph{Proc. ASP-DAC}, pp.~505--512, 2025.
\newblock DOI: \url{10.1145/3658617.3697597}.

\bibitem{khan2025deepseq2}
S.~Khan, Z.~Shi, Z.~Zheng, M.~Li, and Q.~Xu.
\newblock {DeepSeq2}: Enhanced sequential circuit learning with disentangled representations.
\newblock In \emph{Proc. ASP-DAC}, pp.~498--504, 2025.
\newblock DOI: \url{10.1145/3658617.3697594}.

\bibitem{yoon2025paraformer}
J.~Yoon, J.~Lee, D.~Kim, J.~Hur, and S.~Kang.
\newblock {ParaFormer}: A hybrid graph neural network and transformer approach for pre-routing parasitic {RC} prediction.
\newblock In \emph{Proc. ASP-DAC}, pp.~513--519, 2025.
\newblock DOI: \url{10.1145/3658617.3697599}.

\bibitem{he2024opttiming}
G.~He, W.~Ding, Y.~Ye, X.~Cheng, Q.~Song, and P.~Cao.
\newblock An optimization-aware pre-routing timing prediction framework based on heterogeneous graph learning.
\newblock In \emph{Proc. ASP-DAC}, pp.~177--182, 2024.
\newblock DOI: \url{10.1109/ASP-DAC58780.2024.10473937}.

\bibitem{luo2025drcircuitgnn}
Y.~Luo, S.~Li, J.~Tao, K.~G. Thorat, X.~Xie, H.~Peng, N.~Xu, C.~Ding, and S.~Huang.
\newblock {DR-CircuitGNN}: Training acceleration of heterogeneous circuit graph neural network on {GPUs}.
\newblock In \emph{Proc. ACM International Conference on Supercomputing (ICS)}, pp.~221--235, 2025.
\newblock DOI: \url{10.1145/3721145.3734528}.

\end{thebibliography}

\appendix

\section{Survey Methodology and Inclusion Criteria}
\label{app:methodology}

We surveyed papers from EDA venues (DAC, ICCAD, DATE, ASP-DAC, MLCAD, ISPD, IEEE TCAD, ACM TODAES) and graph-learning venues (NeurIPS, ICLR, ICML, AAAI, KDD, WWW) appearing through early May 2026, to the extent that stable public bibliographic metadata were available; very recent works with unstable bibliographic metadata were excluded from primary claims. A paper was included if it satisfied at least one of: (i) it proposes or uses a graph neural architecture for an EDA task; (ii) it defines a graph-structured representation of a circuit and uses it for a learning problem; or (iii) it provides an open benchmark, dataset, or critical assessment of GNN-for-EDA methods. Purely tabular ML predictors that do not define a graph were excluded; non-GNN optimization baselines (e.g., differentiable placers) are listed separately in Table~\ref{tab:eda_apps} as context. We additionally exclude continuous-equivariant geometric GNNs, for the reasons in \S\ref{sec:intro}.

\section{Detailed GNN Architecture Overview}
\label{app:gnn_review}

\paragraph{Spectral methods.} The earliest GNNs derived graph convolution from the Laplacian eigendecomposition~\cite{bruna2014}; ChebNet~\cite{defferrard2016} approximated the spectral filter with Chebyshev polynomials at $O(K|\mathcal{E}|)$ cost; GCN~\cite{kipf2017} simplified to first-order. The spectral story is largely subsumed by spatial methods in EDA practice.

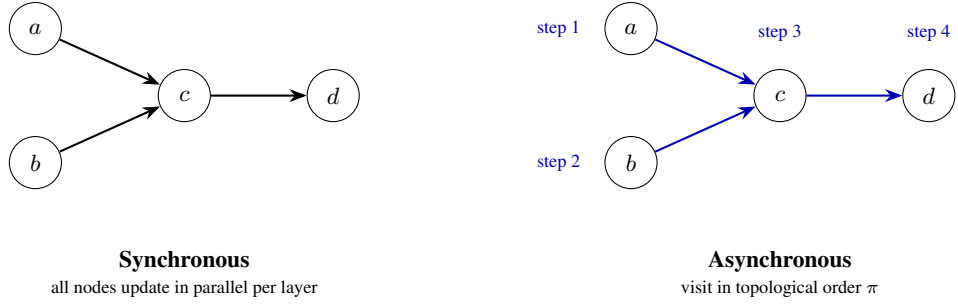
\begin{figure}[h]
\centering
\resizebox{0.9\textwidth}{!}{%
\begin{tikzpicture}[
  node distance=8mm and 12mm,
  every node/.style={font=\small},
  cell/.style={circle,draw,minimum size=7mm,inner sep=0pt},
  arr/.style={->,>=Stealth,thick},
  larr/.style={->,>=Stealth,thick,color=blue!70!black},
  steplab/.style={font=\scriptsize,blue!70!black},
]
\node[cell] (s1) at (0,2.6) {$a$};
\node[cell] (s2) at (0,0.8) {$b$};
\node[cell] (s3) at (2.0,1.7) {$c$};
\node[cell] (s4) at (4.0,1.7) {$d$};
\draw[arr] (s1) -- (s3);
\draw[arr] (s2) -- (s3);
\draw[arr] (s3) -- (s4);
\node[align=center] at (2.0,-0.7) {\textbf{Synchronous}\\\scriptsize all nodes update in parallel per layer};

\begin{scope}[xshift=8.0cm]
\node[cell] (a1) at (0,2.6) {$a$};
\node[cell] (a2) at (0,0.8) {$b$};
\node[cell] (a3) at (2.0,1.7) {$c$};
\node[cell] (a4) at (4.0,1.7) {$d$};
\draw[larr] (a1) -- (a3);
\draw[larr] (a2) -- (a3);
\draw[larr] (a3) -- (a4);
\node[steplab,left=2mm of a1]  {step 1};
\node[steplab,left=2mm of a2]  {step 2};
\node[steplab,above=2.5mm of a3] {step 3};
\node[steplab,above=2.5mm of a4] {step 4};
\node[align=center] at (2.0,-0.7) {\textbf{Asynchronous}\\\scriptsize visit in topological order $\pi$};
\end{scope}
\end{tikzpicture}%
}
\caption{Synchronous vs.\ asynchronous DAG message passing on a small DAG. The asynchronous variant (Eq.~\ref{eq:dag_async}) makes the entire transitive fan-in cone reachable in a single forward pass; step labels indicate the topological visit order $\pi$.}
\label{fig:topo_mpnn}
\end{figure}

\paragraph{Synchronous vs.\ asynchronous DAG message passing.} Figure~\ref{fig:topo_mpnn} contrasts the two regimes. Synchronous propagation is regime-independent; asynchronous propagation aligns with the topological ordering of the underlying recurrence, at the cost of a serialized forward pass; whether the learned aggregation reproduces the exact max/min algebra of the target task is a separate question of operator and capacity.

\paragraph{GAT and GIN equations.} For completeness: GAT computes
\begin{equation*}
\alpha_{ij} = \frac{\exp\!\left(\mathrm{LeakyReLU}(\mathbf{a}^\top [\mathbf{W}\mathbf{h}_i \| \mathbf{W}\mathbf{h}_j])\right)}{\sum_{k \in \mathcal{N}(i)} \exp\!\left(\mathrm{LeakyReLU}(\mathbf{a}^\top [\mathbf{W}\mathbf{h}_i \| \mathbf{W}\mathbf{h}_k])\right)},
\end{equation*}
with GATv2~\cite{brody2022gatv2} swapping the order of LeakyReLU and projection to give dynamic (query-dependent) attention. GIN updates with $\mathbf{h}_v^{(k)} = \mathrm{MLP}^{(k)}\!\big((1+\epsilon^{(k)})\mathbf{h}_v^{(k-1)} + \sum_{u \in \mathcal{N}(v)} \mathbf{h}_u^{(k-1)}\big)$ and matches 1-WL.

\paragraph{HGT typed attention.} For relation $r \in \mathcal{R}$, HGT~\cite{hu2020hgt} computes per-relation $\mathbf{q}_v^r = \mathbf{W}_Q^r \mathbf{h}_v$, $\mathbf{k}_u^r = \mathbf{W}_K^r \mathbf{h}_u$, $\mathbf{v}_u^r = \mathbf{W}_V^r \mathbf{h}_u$, with
$\alpha_{vu}^r = \mathrm{softmax}_u\!\big(\langle \mathbf{q}_v^r, \mathbf{k}_u^r\rangle / \sqrt{d}\big)$
and $\mathbf{h}_v' = \sum_r \sum_{u \in \mathcal{N}_r(v)} \alpha_{vu}^r \mathbf{v}_u^r$.

\paragraph{Hypergraph two-phase aggregation.} HGNN~\cite{feng2019hgnn} uses
$\mathbf{h}_e = \psi(\{\mathbf{h}_v : v \in e\})$ followed by $\mathbf{h}_v' = \phi(\mathbf{h}_v, \{\mathbf{h}_e : v \in e\})$.

\section{Multi-Die Graph Taxonomy}
\label{app:multidie}

A 2.5D / 3D system supports several graph views, which we list explicitly because no single representation captures all relevant constraints:
\begin{itemize}[noitemsep,leftmargin=*]
\item \emph{Die-level graph}: chiplets / dies as super-nodes; interposer routes, micro-bumps, TSV stacks as edges with their own electrical and reliability attributes.
\item \emph{Block-level graph}: chiplet floorplan, thermal domains, and power-domain hierarchy.
\item \emph{Net-level graph}: cross-die timing paths through inter-chiplet connections.
\item \emph{PDN graph}: vertical current paths through TSVs and micro-bumps; couples to thermal.
\item \emph{Thermal graph}: heat-diffusion coupling between stacked dies and across the interposer.
\item \emph{Packaging graph}: substrate / interposer routing resources and reliability constraints.
\end{itemize}
Heterogeneous GNNs that simultaneously reason at within-die and across-die scales remain rare; multi-die EDA flow examples such as Pin-3D~\cite{pentapati2023pin3d} provide a context for future learned approaches.

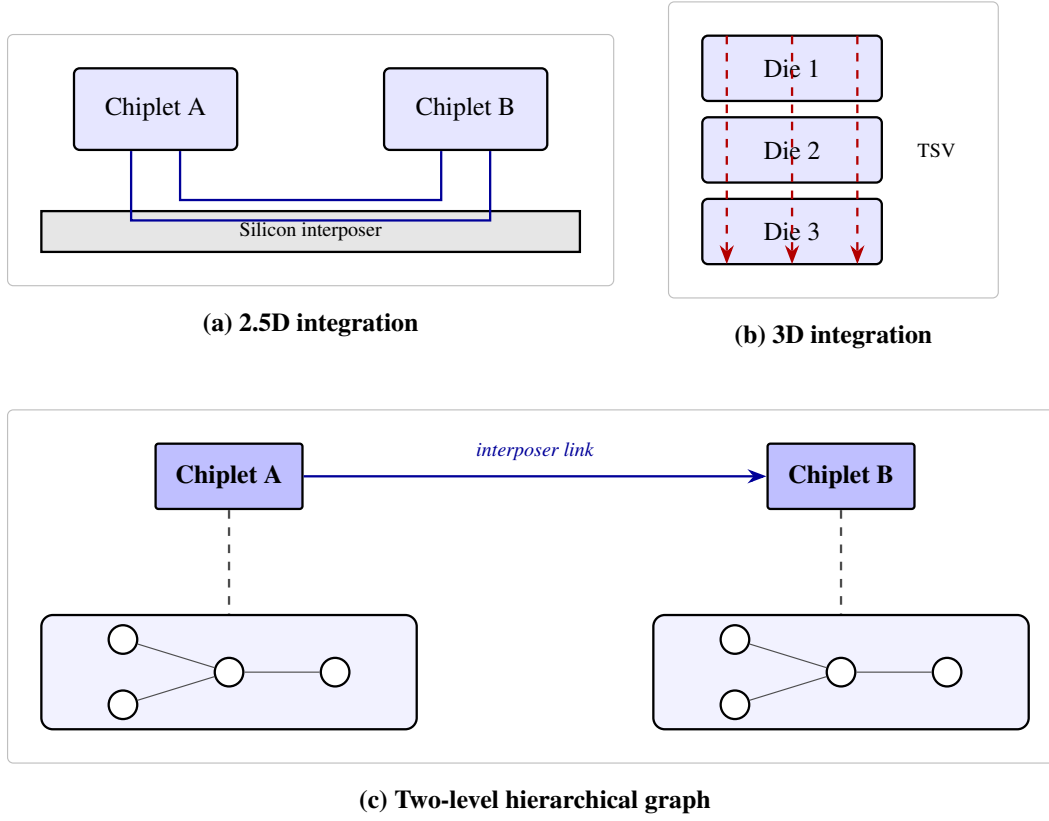
\begin{figure}[h]
\centering
\resizebox{\textwidth}{!}{%
\begin{tikzpicture}[
  every node/.style={font=\small},
  panelframe/.style={draw,thin,gray!50,rounded corners=2pt,inner sep=4mm},
  chiplet/.style={rectangle,draw,thick,rounded corners=2pt,fill=blue!10,minimum width=20mm,minimum height=10mm,align=center,font=\small},
  diebox/.style={rectangle,draw,thick,rounded corners=2pt,fill=blue!10,minimum width=22mm,minimum height=8mm,align=center,font=\small},
  supernode/.style={rectangle,draw,thick,fill=blue!25,rounded corners=1pt,minimum width=18mm,minimum height=8mm,align=center,inner sep=2pt,font=\small\bfseries},
  region/.style={draw,thick,fill=blue!5,rounded corners=4pt},
  innernode/.style={circle,draw,thick,fill=white,minimum size=3.5mm,inner sep=0pt},
  ipos/.style={rectangle,draw,thick,fill=gray!20,minimum width=66mm,minimum height=5mm},
  tsv/.style={->,>=Stealth,thick,red!70!black,dashed},
  hier/.style={thin,gray!60!black},
  contain/.style={thick,gray!60!black,dashed},
  intercon/.style={thick,blue!60!black},
  panel/.style={font=\small\bfseries,align=center},
]
\begin{scope}
\node[chiplet] (CL) at (1.6, 0.5) {Chiplet A};
\node[chiplet] (CR) at (5.4, 0.5) {Chiplet B};
\node[ipos] (interposer) at (3.5,-1.0) {};
\node[font=\scriptsize] at (3.5,-1.0) {Silicon interposer};
\draw[intercon] ($(CL.south)+(3mm,0)$) -- ++(0,-0.6) -- ($(CR.south)+(-3mm,-0.6)$) -- ++(0,0.6);
\draw[intercon] ($(CL.south)+(-3mm,0)$) -- ++(0,-0.85) -- ($(CR.south)+(3mm,-0.85)$) -- ++(0,0.85);
\node[panelframe,fit=(CL)(CR)(interposer)] (Apanel) {};
\node[panel,below=2mm of Apanel] {(a) 2.5D integration};
\end{scope}

\begin{scope}[xshift=9.4cm]
\node[diebox] (D1) at (0, 1.0) {Die 1};
\node[diebox] (D2) at (0, 0.0) {Die 2};
\node[diebox] (D3) at (0,-1.0) {Die 3};
\foreach \x in {-0.8,0.0,0.8} { \draw[tsv] (\x,1.4) -- (\x,-1.4); }
\node[font=\scriptsize,right=3mm of D2] (tsvLab) {TSV};
\node[panelframe,fit=(D1)(D3)(tsvLab)] (Bpanel) {};
\node[panel,below=2mm of Bpanel] {(b) 3D integration};
\end{scope}

\begin{scope}[shift={(0,-5.0)}]
  \node[supernode] (sA) at (2.5, 1.0) {Chiplet A};
  \node[supernode] (sB) at (10.0, 1.0) {Chiplet B};
  \draw[intercon,->,>=Stealth,thick] (sA) -- (sB)
    node[midway,above=0.7mm,font=\scriptsize\itshape] {interposer link};

  \node[region,minimum width=46mm,minimum height=14mm] (regA) at (2.5,-1.4) {};
  \node[region,minimum width=46mm,minimum height=14mm] (regB) at (10.0,-1.4) {};

  \node[innernode] (a1) at (1.2,-1.0) {};
  \node[innernode] (a2) at (1.2,-1.8) {};
  \node[innernode] (a3) at (2.5,-1.4) {};
  \node[innernode] (a4) at (3.8,-1.4) {};
  \draw[hier] (a1) -- (a3); \draw[hier] (a2) -- (a3); \draw[hier] (a3) -- (a4);

  \node[innernode] (b1) at (8.7,-1.0) {};
  \node[innernode] (b2) at (8.7,-1.8) {};
  \node[innernode] (b3) at (10.0,-1.4) {};
  \node[innernode] (b4) at (11.3,-1.4) {};
  \draw[hier] (b1) -- (b3); \draw[hier] (b2) -- (b3); \draw[hier] (b3) -- (b4);

  \draw[contain] (sA) -- (regA.north);
  \draw[contain] (sB) -- (regB.north);


  \node[panelframe,fit=(sA)(sB)(regA)(regB)] (Cpanel) {};
  \node[panel,below=2mm of Cpanel] {(c) Two-level hierarchical graph};
\end{scope}
\end{tikzpicture}%
}
\caption{Multi-die systems as hierarchical graphs. (a)~2.5D integration: chiplets share a silicon interposer with high-density inter-chiplet routing. (b)~3D integration: dies are stacked vertically with through-silicon vias (TSVs). (c)~Both map to a two-level hierarchical graph in which chiplet super-nodes \emph{contain} within-die netlist sub-graphs (shaded regions, dashed containment edges); heterogeneous GNNs can in principle reason at both scales simultaneously.}
\label{fig:multi_die}
\end{figure}

\section{Digital Design Flow Reference Map}
\label{app:designflow}

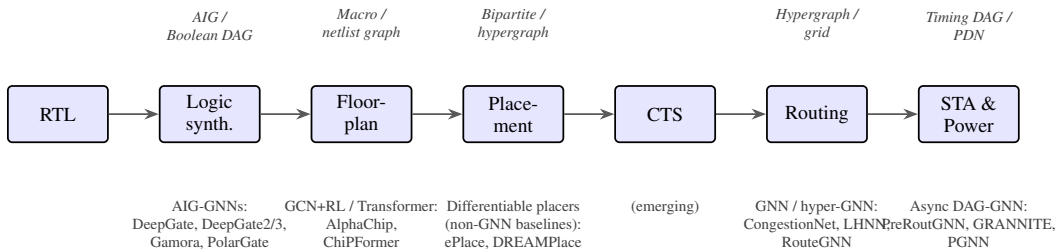
\begin{figure}[h]
\centering
\resizebox{\textwidth}{!}{%
\begin{tikzpicture}[
  node distance=8mm and 8mm,
  every node/.style={font=\small},
  stage/.style={rectangle,draw,thick,rounded corners=2pt,fill=blue!10,minimum width=16mm,minimum height=9mm,align=center,inner sep=2pt},
  arr/.style={->,>=Stealth,thick,gray!70!black},
  toplabel/.style={font=\scriptsize\itshape,align=center,gray!50!black},
  botlabel/.style={font=\scriptsize,align=center,gray!50!black},
]
\node[stage] (rtl)   {RTL};
\node[stage,right=of rtl]   (syn)   {Logic\\synth.};
\node[stage,right=of syn]   (fp)    {Floor-\\plan};
\node[stage,right=of fp]    (place) {Place-\\ment};
\node[stage,right=of place] (cts)   {CTS};
\node[stage,right=of cts]   (route) {Routing};
\node[stage,right=of route] (sta)   {STA \&\\Power};
\draw[arr] (rtl) -- (syn);
\draw[arr] (syn) -- (fp);
\draw[arr] (fp)  -- (place);
\draw[arr] (place) -- (cts);
\draw[arr] (cts) -- (route);
\draw[arr] (route) -- (sta);
\node[botlabel,below=8mm of syn]   {AIG-GNNs:\\DeepGate, DeepGate2/3,\\Gamora, PolarGate};
\node[botlabel,below=8mm of fp]    {GCN+RL / Transformer:\\AlphaChip,\\ChiPFormer};
\node[botlabel,below=8mm of place] {Differentiable placers\\(non-GNN baselines):\\ePlace, DREAMPlace};
\node[botlabel,below=8mm of cts]   {(emerging)};
\node[botlabel,below=8mm of route] {GNN / hyper-GNN:\\CongestionNet, LHNN,\\RouteGNN};
\node[botlabel,below=8mm of sta]   {Async DAG-GNN:\\PreRoutGNN, GRANNITE,\\PGNN};
\node[toplabel,above=6mm of syn]   {AIG /\\Boolean DAG};
\node[toplabel,above=6mm of fp]    {Macro /\\netlist graph};
\node[toplabel,above=6mm of place] {Bipartite /\\hypergraph};
\node[toplabel,above=6mm of route] {Hypergraph /\\grid};
\node[toplabel,above=6mm of sta]   {Timing DAG /\\PDN};
\end{tikzpicture}%
}
\caption{The digital design flow with representative GNN methods (and non-GNN baselines) at each stage. Italics above each stage indicate the dominant graph representation; lists below each stage give representative methods discussed in the body.}
\label{fig:design_flow}
\end{figure}

\FloatBarrier


\end{document}